\documentclass[12pt]{article}
\usepackage[margin=1in]{geometry} 
\usepackage{graphicx} 
\usepackage{amsmath}
\usepackage{listings}
\usepackage{authblk}

\usepackage{hyperref}
\renewcommand\Affilfont{\small}
\renewcommand\Authands{ and }

\newcommand{\email}[1]{\href{mailto:#1}{\nolinkurl{#1}}}

\author[1,2]{Nassim Dehouche\thanks{\email{nassim.deh@mahidol.edu}, \email{ndehouche@haai.info}}}
\author[3]{Daniel Friedman \thanks{\email{daniel@activeinference.institute}}}
\affil[1]{Mahidol University International College, Salaya 73170, Thailand}
\affil[2]{HaAI Labs, Créteil 94000, France}
\affil[3]{Active Inference Institute, Davis, California 95616, USA}

\title{Enhancing Population-based Search with Active Inference}

\date{August 2024}

\begin{document}

\maketitle

\begin{abstract}
The Active Inference framework models perception and action as a unified process, where agents use probabilistic models to predict and actively minimize sensory discrepancies. In complement and contrast, traditional population-based metaheuristics rely on reactive environmental interactions without anticipatory adaptation. This paper proposes the integration of Active Inference into these metaheuristics to enhance performance through anticipatory environmental adaptation. We demonstrate this approach specifically with Ant Colony Optimization (ACO) on the Travelling Salesman Problem (TSP). Experimental results indicate that Active Inference can yield some improved solutions with only a marginal increase in computational cost, with interesting patterns of performance that relate to number and topology of nodes in the graph. Further work will characterize where and when different types of Active Inference augmentation of population metaheuristics may be efficacious.
\end{abstract}

\section{Introduction and Basic Concepts}

\subsection{Population-based Metaheuristics}
Population-based metaheuristics (PBMH) \cite{Liu2011}, such as Ant Colony Optimization (ACO), Genetic Algorithms (GA), Particle Swarm Optimization (PSO), and Differential Evolution (DE),  represent a broad class of optimization algorithms that utilize a population of potential solutions to explore and exploit search spaces effectively. These algorithms draw inspiration from various natural and social processes, and they have been widely used to solve complex optimization problems across numerous domains. PBMH are characterized by iterative processes, population-based structures, guided random searches, and parallel processing capabilities. They often embody principles from biology, social behavior, or other natural phenomena. However, they typically employ agents that react passively to predefined rules and environmental data. These approaches generally lack a mechanism for predicting changes or actively shaping their environment, focusing instead on optimizing solutions based on reactive interactions with their surroundings.

\subsection{The Travelling Salesman Problem}
The Travelling Salesman Problem (TSP) is a combinatorial optimization problem defined on a graph 
$G=(V,E)$, where $V$ represents a set of vertices (cities) and $E$ represents a set of edges (paths between cities). The objective is to determine the shortest possible Hamiltonian cycle (tour) that visits each vertex (city)  exactly once and returns to the starting vertex (city) \cite{Applegate2006}. The Travelling Salesman Problem (TSP) is a classic NP-hard optimization problem that plays a crucial role in the P vs NP question. The decision version of TSP, which asks whether a tour of a given length exists, is NP-complete, meaning it is in NP and at least as hard as any problem in NP. 

\subsection{Active Inference}
The framework of Active Inference presents a compelling approach for understanding perception and action across systems and scales \cite{Friston2022, Parr2022}. Cognitive systems such as the brain are represented with probabilistic models that predict and minimize sensory input discrepancies, effectively bounding and reducing uncertainty (minimization of surprise is equivalent to maximization of model evidence). Active inference generative models can integrate perception, action, and learning into a unified framework, capable of considering ecosystems of shared intelligence in terms of active shaping of sensory inputs (rather than passive receipt).\\ 

Active Inference has been applied across several application domains. In psychiatry, active inference models are used to understand and treat disorders such as schizophrenia and depression. These models help explain the altered perception and cognitive biases observed in these conditions, providing insights into more effective therapeutic interventions \cite{Friston2016}. Cognitive science applies active inference to study processes such as attention, memory, and learning, helping to develop computational models that mimic human cognitive functions and improve artificial intelligence systems \cite{Hohwy2020}. In economics, active inference is utilized in models to predict market behaviors and decision-making processes, aiding in understanding how agents optimize their actions based on predictions of future states \cite{Smith2013}. The field of robotics implements active inference to enhance autonomous decision-making and adaptability in dynamic environments. Robots can predict the consequences of their actions and adjust their behavior accordingly \cite{Lanillos2021}.\\

In computational applications, active inference has been increasingly applied to problems such as optimization and machine learning. By integrating predictive coding and Bayesian inference, algorithms can adaptively update their parameters to improve performance on tasks like classification, regression, and clustering \cite{Millidge2021}. In reinforcement learning, active inference frameworks allow agents to not only learn from their actions but also anticipate and plan for future states, enhancing decision-making in complex environments \cite{Tschantz2020}. Active inference has also been applied to the multi-agent setting (e.g. in human organizations and ant colonies), a setting which is extended in this work to the context of population-based search algorithms. 

\subsection{Objectives}

This research explores the performance and computational resourcing associated with using Active Inference to augment population-based metaheuristics (PBMH), in this case the Ant Colony Optimization (ACO) approach to the Travelling Salesman Problem (TSP).\\ 

More generally we seek to consider Active Inference as an alternative or supplemental strategy for traditionally reactive agent-based models and population-type metaheuristics.  \\

This manuscript is structured as follows.  Following this introductory section, Section 2 recalls  the core principles of  population-based metaheuristics (PBMH), including social-cooperation, self-adaptation, and competition, and how these components are instantiated in different PBMH search algorithms. Section 3 situates our integration of active inference into this framework and instantiates the proposed approach on Ant Colony Optimization for the travelling Salesman Problems. Section 4 outlines the experimental setup used to evaluate the performance of the proposed approach and summarizes and discusses our findings. Finally, Section 5 summarizes the  potential significance of our results in enhancing PBMH and suggests directions for future research.

\section{Population-Based Metaheuristics (PBMH)}

The core philosophy behind PBMH search is the collaborative and competitive interplay among a population of agents. Operationally, a PBMH involves a series of procedures that can adaptive balance the exploration of new solutions, with the exploitation of known good solutions. Following \cite{Liu2011}, the main elements of this collective process can be described with three primary components: social-cooperation, self-adaptation, and competition.

\begin{itemize}
    \item \textbf{Social-Cooperation:} This component represents the collaborative effect among individual agents in the population, enabling them to exchange information and learn from each other. For instance, in Particle Swarm Optimization (PSO), particles adjust their positions based on both their personal best positions and the global best positions found by the swarm \cite{Kennedy2001}. Other algorithms may involve direct communication among agents, or stigmergic coordination among agents mediated through the environment. 
    \item \textbf{Self-Adaptation:} Self-adaptation involves each individual adjusting itself independently based on its own experience and surroundings. This process helps in intensifying the search around promising areas while diversifying the search to avoid premature convergence. For instance, in Genetic Algorithms (GA), mutation serves as a self-adaptation mechanism that introduces genetic diversity \cite{Holland1975}.
    \item \textbf{Competition:} This component deals with the selection and updating of the population for the next iteration. Competitive selection ensures that better solutions are retained, thus guiding the population towards optimal solutions. For instance, in Differential Evolution (DE), the selection mechanism compares offspring with their parents and retains the better solutions \cite{Price2005}.
\end{itemize}

\subsection{Examples of Biology-inspired PBMH Algorithms}
\subsubsection{Particle Swarm Optimization (PSO)}
PSO is inspired the collective behavior activities such as bird flocking and fish schooling, processes that operate without stigmergic (niche modification-based) interaction among agents. PSO adjusts the trajectories of individual solutions (particles) based on their own experience and that of their neighbors. For the TSP, modifications are often necessary to maintain the feasibility of particles as they converge towards a solution \cite{goldberg1985}. Each particle in PSO adjusts its trajectory based on its velocity, personal best position and the global best position found by the swarm. The velocity update equation for a particle is given by:
\begin{equation}
    v_i(t+1) = wv_i(t) + c_1r_1(p_i(t) - x_i(t)) + c_2r_2(p_g(t) - x_i(t))
\end{equation}
where $v_i$ is the velocity of particle $i$, $w$ is the inertia weight, $c_1$ and $c_2$ are acceleration coefficients, $r_1$ and $r_2$ are random numbers, $p_i$ is the personal best position of particle $i$, and $p_g$ is the global best position \cite{Kennedy2001}.

\subsubsection{Differential Evolution (DE)}
DE utilizes differential operators to create new candidate solutions and employs a one-to-one competition scheme to select better solutions. The mutation operation in DE involves adding the weighted difference between two randomly selected individuals to a third individual, a process analogous to reproduction with diversification and inter-generational evolution (drift-selection):
\begin{equation}
    v_i = x_{best} + F(x_{r1} - x_{r2})
\end{equation}
where $v_i$ is the mutant vector, $x_{best}$ is the best individual in the current population, $x_{r1}$ and $x_{r2}$ are two randomly selected individuals, and $F$ is the scaling factor \cite{Price2005}.

\subsubsection{Genetic Algorithms (GA)}
GA are even more directly inspired by the principles of natural selection and population genetics. GA implement motifs  such as population selection, meiotic crossover, and mutation operators to generate and evolve populations of solutions. The selection operator chooses parents based on their fitness, the crossover operator combines parents to produce offspring, and the mutation operator introduces genetic diversity. The basic GA process is summarized as:
\begin{equation}
    \text{GA} = (\text{selection}, \text{crossover}, \text{mutation}, \text{fitness})
\end{equation}
 The adaptability of GA is beneficial for exploring varied graph structures, allowing them to maintain diversity in the solution pool and avoid local optima \cite{holland1992}.
 
\subsubsection{Ant Colony Optimization (ACO)}
Ant Colony Optimization (ACO), inspired by the stigmergic foraging behavior of ants, utilizes a collective learning process where simulated ants construct solutions and deposit pheromones to guide subsequent ants towards promising edges. The pheromone concentration influences the probability of selecting a particular edge in subsequent iterations. In the reduced case where there is a single pheromone with simple positive behavior attractivity, the pheromone update rule is given by:
\begin{equation}
    \tau_{ij} = (1 - \rho)\tau_{ij} + \Delta \tau_{ij}
\end{equation}
where $\tau_{ij}$ is the pheromone concentration on edge $(i, j)$, $\rho$ is the evaporation rate, and $\Delta \tau_{ij}$ is the pheromone deposited by ants \cite{Dorigo2004}.
In the context of the Travelling Salesman Problem, ACO has been adapted to favor tours that cover all vertices without closing the loop prematurely. Studies have shown that ACO can effectively approximate Hamiltonian cycles, particularly in sparse graphs where the pheromone trails help to highlight potential pathways through less connected vertices \cite{yu2022}.

\section{Proposed Approach}
This study proposes incorporating principles of Active Inference to enhance the ACO approach to TSP, more generally aligning with the core components of PBMH: social-cooperation, self-adaptation, and competition with predictive and adaptive mechanisms. These enhancements are initial developments towards a broader program of allowing development-ecological-evolutionary cognitive agents to not only react to the environment but also anticipate future states and adjust their strategies accordingly. Our algorithm introduces a belief update mechanism and free energy calculation, enabling adaptive decision-making based on the current tour quality. The proposed method consists of the following components and processes:

\subsection{Active Inference augmented ACO Algorithm applied to TSP}
\begin{enumerate}
\item \textbf{Initialization:}
The algorithm initializes with (N) nodes from the input graph, sets uniform initial pheromone levels across all edges, and establishes baseline values for the best tour and its length.
\item \textbf{Tour Length Calculation:}
A function \texttt{calculate\_path\_length(path)} computes the total length of a given tour, accounting for the return to the starting node.
\item \textbf{Node Selection:}
The function \texttt{choose\_next\_node(available\_nodes, current\_node, belief)} selects subsequent nodes based on pheromone levels, heuristic information, and the current belief state. The probability of selecting node \(j\) from node \(i\) is calculated as:
\begin{equation}
    p_{ij} = \frac{(\tau_{ij})^\alpha \cdot (1/d_{ij})^\beta \cdot belief}{\underset{{k \in available}}\sum (\tau_{ik})^\alpha \cdot (1/d_{ik})^\beta \cdot belief}
\end{equation}
where \(\tau_{ij}\) represents the pheromone level, \(d_{ij}\) the distance, and \(\alpha\) and \(\beta\) are parameters controlling the influence of pheromone and distance respectively.
\item \textbf{Free Energy Calculation:}
The function \texttt{free\_energy(belief\_in\_tour, path\_length)} computes the free energy for a given tour, combining the expected path length and associated uncertainty:
\begin{equation}
    F = path\_length - (b \log b + (1-b) \log (1-b))
\end{equation}
where \(b\) represents the belief in the current tour.
\item \textbf{Iterative Optimization Process:}
The main algorithm iterates for a specified number of cycles. Each iteration comprises:
\begin{itemize}
    \item Tour construction by individual ants, starting from random nodes.
    \item Dynamic belief updating based on the current tour length relative to the best known tour:
    \begin{equation}
        belief = 1 - \frac{current\_path\_length}{best\_path\_length}
    \end{equation}
    \item Pheromone update with evaporation and deposition:
    \begin{equation}
        \tau_{ij} = (1 - \rho)\tau_{ij} + \frac{\Delta \tau}{path\_length}
    \end{equation}
    where \(\rho\) denotes the evaporation rate and \(\Delta \tau\) the pheromone deposit.
    \item Implementation of an elitist strategy, reinforcing the best tour:
    \begin{equation}
        \tau_{ij}^{best} = \tau_{ij}^{best} + \frac{2 \Delta \tau}{best\_path\_length}
    \end{equation}
\end{itemize}

\item \textbf{Output:}
The algorithm outputs the optimal tour discovered and its corresponding length.

\end{enumerate}

This enhanced ACO algorithm integrates active inference principles through the belief update mechanism and free energy calculation. The pheromone distribution represents colony-scale adaptive beliefs about the distribution of tours. Nestmate-level node selection based upon phermone density, allows for dynamic agent-scale, tour-specific, adjustment of behavior. The possibility to then employ single heuristics (e.g. elitist or fitness-based strategies where the impact of high-quality solutions is amplified, potentially accelerating convergence to optimal or near-optimal tours).

\section{Experiments}
\subsection{Experimental Setup}
To evaluate the performance of our Active Inference-enhanced Ant Colony Optimization (AI ACO) algorithm, we conducted a series of experiments comparing it against a basic ACO, and a baseline Nearest Neighbor heuristic (NN). The focus of these experiments was to assess relative performance improvements rather than absolute benchmarking against other methods. We performed the comparison on randomly generated symmetric graphs with varying numbers of nodes to simulate different problem complexities. Specifically, we generated 100 random graphs for each of the following sizes: 50, 100, 250, and 500 nodes. Each graph was constructed to be symmetric with no self-loops. Experimental results are presented in Table \ref{tab:results}.\\

Python implementations of the variaous methods considered in our experiments are provided in the Appendix of this manuscript. Extended statistical results, including a detailed analysis of variance (ANOVA) can be found in the following Github repository \url{https://github.com/haailabs/ActiveACO}.

\subsection{Results}

\begin{table}[h]
\centering

\caption{Comparison of Basic ACO and Active Inference-enhanced ACO (50 graphs per node size)}
\label{tab:results}
\begin{tabular}{lrrrrr}
\hline
\textbf{Metric} & \textbf{25 Nodes} & \textbf{50 Nodes} & \textbf{100 Nodes} & \textbf{250 Nodes} & \textbf{500 Nodes}\\
\hline
NN Tour Length (mean) & 24.667 & 33.255 & 45.151 & 83.732 &136.489 \\
NN Tour Length (std) & 17.978 & 40.995& 54.799 & 149.660 &226.256\\
ACO Tour Length (mean) & 8.359 & 16.984 & 22.455 & 60.552 &108.240\\
ACO Tour Length (std) & 10.883 & 34.470 & 21.736 & 109.039 &77.320\\
AI ACO Tour Length (mean) & 8.149 & 16.368 & 20.477 & 56.587 &100.350
\\
AI ACO Tour Length (std) & 10.853 & 33.358 & 20.745 & 110.691 &75.368
\\
Improvement \% (mean) & 2.52\%	&3.62\%	&8.81\%	&6.55\% &7.29\%\\
NN Time s (mean) & 0.0001 & 0.0002 & 0.005 & 0.0132 &0.029 \\
NN Time s (std) & 0.0000 & 0.0019 & 0.0064 & 0.0053 & 0.012\\
ACO Time s (mean) & 0.814 & 2.099 & 6.076 & 29.229&163.893 \\
ACO Time s (std) & 0.017 & 0.015 & 0.071 & 0.221 &38.950
\\
AI ACO Time s (mean) & 0.891 & 2.261 & 6.382 & 29.973 &167.120
\\
AI ACO Time s (std) & 0.010 & 0.034 & 0.040 & 0.134 &13.074\\
Time Increase s  & 0.077	&0.162&	0.306&	0.744	&3.227\\
Time Increase \%  & 9.46\%	&7.72\%	&5.04\%	&2.55\%	&1.97\%\\
\hline
\end{tabular}
  \label{results}
\end{table}

\subsubsection{Performance Improvement}
The Active Inference-enhanced ACO demonstrated consistent improvement in tour length across all graph sizes tested. For 25-node graphs, the average improvement was 2.52\%. This improvement increased to 3.62\% for 50-node graphs and rose significantly to 8.81\% for 100-node graphs. For larger graphs, the improvement remained substantial. The 250-node graphs showed a 6.55\% improvement, while 500-node graphs exhibited a 7.29\% improvement. This trend suggests that the Active Inference enhancement continues to offer significant benefits for larger problem sizes, with the rate of improvement stabilizing for very large instances.

It is important to note the high variability in tour lengths for both algorithms, as indicated by the large standard deviations. For instance, in 500-node graphs, the standard deviation for the Basic ACO is 77.320, while for the AI-enhanced ACO it is 75.368. The variability does not show a consistent trend with increasing graph size, as seen in the fluctuations of standard deviations across different node counts. This suggests that the performance of both algorithms can be less predictable for larger, more complex problems, but the relationship between problem size and variability is not straightforward.

\subsubsection{Computation Time}
Computation times unsurprisingly increased with graph size. The Basic ACO times ranged from 0.814813 seconds for 25-node graphs to 163.893 seconds for 500-node graphs. Similarly, the AI-enhanced ACO times ranged from 0.891167 seconds for 25-node graphs to 167.120 seconds for 500-node graphs. However, relative to the total computation times, the percentages of increase in computation time are decreasing, from 9.46\% seconds for 25-node graphs to 1.97\% seconds for 500-node graphs. Interestingly, the standard deviation of computation times for the AI-enhanced ACO was generally smaller than for the Basic ACO, particularly for larger graphs. For instance, in 500-node graphs, the standard deviation for the Basic ACO is 38.950 seconds, while for the AI-enhanced ACO it is 13.074 seconds. This suggests that the enhanced version offers more consistent performance in terms of computation time for larger problem instances.

\subsubsection{Scalability and Efficiency}
The previous results indicate that the Active Inference-enhanced ACO scales well with increasing graph size. The improvement in tour length shows increased effectiveness for more complex problems. Concurrently, the relative increase in computation time decrease from 9.46\% seconds for 25-node graphs to 1.97\% seconds for 500-node graphs. 

In absolute terms, the additional computation time required by the AI-enhanced ACO increased with problem size, from 0.077 seconds for 25-node graphs to 3.227 seconds for 500-node graphs. However, the relative computational overhead decreased as the problem size grew larger.

\begin{figure}[h]
    \centering
    \includegraphics[width=.8\textwidth]{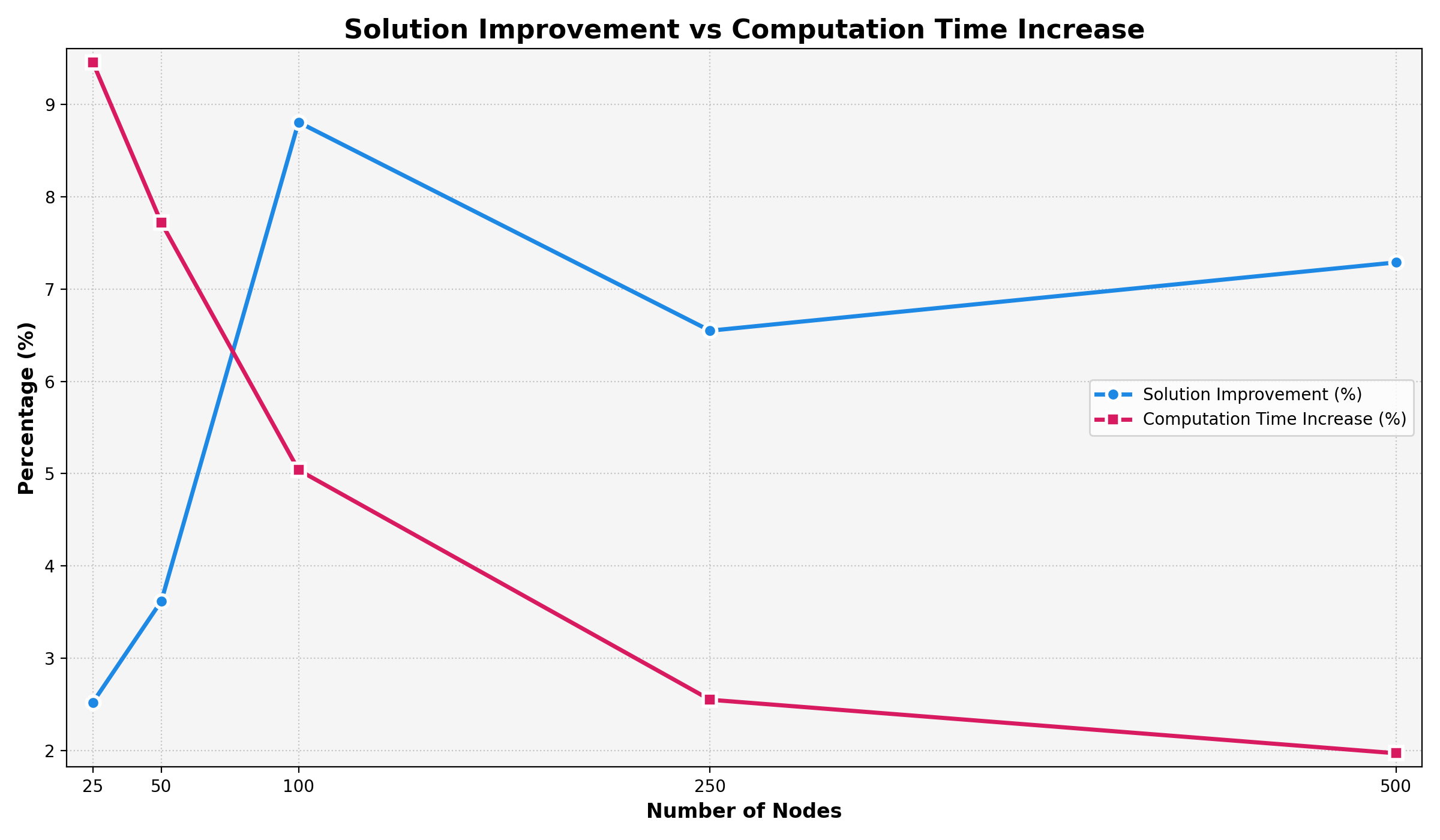}
    \caption{Scaling of solution quality improvement and computational overhead with problem size.}
    \label{fig:scale}
\end{figure}

This trend indicates that the Active Inference enhancement becomes more efficient in terms of the trade-off between solution quality improvement and computational cost as the problem size increases. For larger problem instances, the algorithm achieves substantial improvements in solution quality with a proportionally smaller increase in computation time, as illustrated in Figure \ref{fig:scale}.

However, it is important to note the high variability in tour lengths for both algorithms, as indicated by the large standard deviations provided in Table \ref{results}. This variability suggests that while the AI-enhanced ACO offers consistent improvements on average, its performance may fluctuate across different problem instances, particularly for larger, more complex problems. Further investigation into the factors contributing to this variability could provide valuable insights for refining the algorithm and predicting when it will be most effective. Additionally, research into methods to reduce this variability while maintaining or improving the average performance could be a fruitful avenue for future work.

\subsubsection{Statistical Analysis}
To assess the performance difference between the traditional ACO and the Active Inference-enhanced ACO, we conducted several statistical tests. These initial statistical results presented here can and will be expanded, enabling better-powered and more comprehensive analyses across types and sizes of graph.  
\begin{itemize}
\item Paired T-test for Tour Length:
T-statistic = 2.7852, P-value = 0.00641\\
This test shows a statistically significant difference in tour lengths between the two algorithms (p $<$ 0.01). The positive T-statistic indicates that the Active Inference-enhanced ACO consistently produces shorter tours compared to the traditional ACO across the tested instances.
\item Wilcoxon Signed-Rank Test:
W-statistic = 217.0, P-value = 0.02158\\
This non-parametric test also reveals a statistically significant difference (p $<$  0.05) in performance between the two algorithms. It confirms the findings of the paired T-test, suggesting that the improvement in tour length by the Active Inference-enhanced ACO is consistent and not due to chance or outliers.
\item Mann-Whitney U Test:
U-statistic = 5291.0, P-value = 0.47783\\
Interestingly, this test does not show a statistically significant difference between the two algorithms. This suggests that while the Active Inference-enhanced ACO consistently outperforms the traditional ACO on the same problem instances (as shown by the paired tests), the overall distributions of tour lengths for the two algorithms across different problem instances are not significantly different.

\item Paired T-test for Computation Time:
T-statistic = -1.7753, P-value = 0.07892\\
This test indicates that there is no statistically significant difference in computation time between the two algorithms at the conventional 0.05 significance level. However, the low P-value ($<$ 0.1) suggests a trend towards the Active Inference-enhanced ACO taking slightly longer to compute, which aligns with our observations from the raw data.
\end{itemize}
These statistical results provide valuable insights into the performance characteristics of the Active Inference-enhanced ACO:

On average, The algorithm produces better solutions (shorter tours) compared to the traditional ACO when solving the same problem instances, as evidenced by the significant results in the paired tests.
However, the lack of significance in the Mann-Whitney U test suggests that this improvement, while consistent, may not be large enough to create a significant difference in the overall distribution of tour lengths across different problem instances.
The computational overhead of the Active Inference enhancement, while observable in the raw data, is not statistically significant at the conventional 0.05 level. This suggests that the improvement in solution quality comes with only a marginal increase in computational cost.

These findings reinforce our earlier observations about the trade-off between solution quality and computation time. The Active Inference-enhanced ACO offers statistically significant improvements in solution quality with only a slight, statistically insignificant increase in computation time. This makes it a promising approach for problems where solution quality is the primary concern and small increases in computation time are acceptable.

\subsection{Discussion}

Functionally, and as demonstrated in the experiments/results, the population of Active Inference ACO solutions appears to harbor some or many slightly-longer tours (reflected by lower median score), at the up
The effectiveness is demonstrated through instantiating ACO for the TSP, where it can improve average performance. These results present a nuanced picture of the Active Inference-enhanced ACO’s performance.
While the algorithm shows promising improvements, particularly for larger graphs, the benefits are not uniform across all instances.
The increasing improvement percentages for different graph sizes suggest that the Active Inference enhancement becomes more effective as
problem complexity increases. However, the growing standard deviations in improvement
percentages indicate that this effectiveness is highly variable and likely depends on specific
graph characteristics.\\

The computational cost of the Active Inference enhancement appears to be relatively minor
for smaller graphs but is projected to increase for larger problems. This trade-off between
solution quality and computation time becomes more pronounced as graph size increases,
and may be an important consideration for practical applications.
The statistical analysis reveals a significant improvement in tour length according to the
paired T-test, but the lack of significance in non-parametric tests suggests that this improvement may not be consistent across all graph instances. This inconsistency highlights
the complexity of the traveling salesman problem and the challenge of developing universally
effective heuristics.
Future research should focus on identifying the specific graph characteristics that lead to
larger improvements with the Active Inference enhancement. Additionally, investigating
the algorithm’s performance on a wider range of graph sizes and structures could provide
valuable insights into its scalability and applicability to different problem domains

\section{Conclusion}
This paper introduced a novel framework that integrates Active Inference with population-based metaheuristics (PBMH), aiming to transform traditionally reactive algorithms into more proactive, adaptive systems. We demonstrated this framework's application to Ant Colony Optimization (ACO) for solving the Travelling Salesman Problem (TSP), with results showing a nuanced improvement in performance.
Our experimental results, based on randomly generated graphs, indicate that the Active Inference-enhanced ACO achieves average improvements in tour length compared to the traditional ACO. These improvements are statistically significant according to both the paired t-test and Wilcoxon Signed-Rank Test. However, the performance gain varies considerably across different graph types, suggesting the need for a more comprehensive understanding of the relationship between problem characteristics and algorithm performance.\\

While our results are promising, several limitations of this study point to important directions for future research. Our focus on ACO and TSP limits the generalizability of our findings. Future work should explore the application of this framework to other PBMHs, such as Genetic Algorithms (GA), Particle Swarm Optimization (PSO), and Differential Evolution (DE). This would provide insights into how the benefits and challenges observed in ACO translate to different algorithms and problem domains. Moreover, further research is needed to develop a stronger theoretical justification for the integration of Active Inference with PBMH. This could help explain the observed variability in performance gains and guide future algorithmic improvements. Given the variability in performance improvement across different graph types, future work could focus on developing adaptive mechanisms that dynamically adjust the influence of Active Inference based on the characteristics of the problem being solved. A more comprehensive analysis of how different graph properties affect the algorithm's performance could provide valuable insights for tailoring the approach to specific problem instances.\\

The integration of Active Inference with PBMH provides a foundation for developing more sophisticated hybrid algorithms. These could potentially leverage the strengths of multiple metaheuristics while using Active Inference to guide the selection and application of different strategies. While our Active Inference-enhanced ACO shows promise in improving performance on the TSP, the results highlight the complexity of enhancing metaheuristics with cognitive-inspired approaches. This work opens up new avenues for research in adaptive optimization algorithms, with the potential to develop more flexible and efficient problem-solving techniques across a wide range of domains.

\appendix
\section*{Appendix: Algorithms Implementation}
\begin{lstlisting}[breakatwhitespace=true, breaklines=true,basicstyle=\ttfamily]

import numpy as np
import random
import pandas as pd
import matplotlib.pyplot as plt
from scipy import stats
from scipy.stats import spearmanr


def ant_colony_optimization(graph, num_ants=10, num_iterations=100, alpha=1.0, beta=2.0, evaporation_rate=0.5, pheromone_deposit=1.0):
    N = graph.shape[0]
    pheromone = np.ones((N, N))
    best_path = None
    best_path_length = float('inf')

    def calculate_path_length(path):
        return sum(graph[path[i], path[i + 1]] for i in range(len(path) - 1)) + graph[path[-1], path[0]]

    def choose_next_node(available_nodes, current_node):
        probabilities = [(pheromone[current_node, node] ** alpha) * ((1.0 / graph[current_node, node]) ** beta) for node in available_nodes]
        probabilities = np.array(probabilities)
        probabilities /= probabilities.sum()
        return np.random.choice(available_nodes, p=probabilities)

    for _ in range(num_iterations):
        all_paths = []
        all_lengths = []
        for _ in range(num_ants):
            path = [random.randint(0, N - 1)]
            available_nodes = list(set(range(N)) - set(path))
            while available_nodes:
                next_node = choose_next_node(available_nodes, path[-1])
                path.append(next_node)
                available_nodes.remove(next_node)
            path_length = calculate_path_length(path)
            all_paths.append(path)
            all_lengths.append(path_length)
            if path_length < best_path_length:
                best_path = path
                best_path_length = path_length
        pheromone *= (1 - evaporation_rate)
        for path, length in zip(all_paths, all_lengths):
            for i in range(len(path) - 1):
                pheromone[path[i], path[i + 1]] += pheromone_deposit / length
            pheromone[path[-1], path[0]] += pheromone_deposit / length

    return best_path, best_path_length

def improved_aco_active_inference(graph, num_ants=10, num_iterations=100, alpha=1.0, beta=2.0, evaporation_rate=0.5, pheromone_deposit=1.0):
    N = graph.shape[0]
    pheromone = np.ones((N, N))
    best_path = None
    best_path_length = float('inf')

    def calculate_path_length(path):
        return sum(graph[path[i], path[i + 1]] for i in range(len(path) - 1)) + graph[path[-1], path[0]]

    def choose_next_node(available_nodes, current_node, belief):
        probabilities = [(pheromone[current_node, node] ** alpha) * ((1.0 / graph[current_node, node]) ** beta) for node in available_nodes]
        probabilities = np.array(probabilities)
        probabilities *= belief  # Adjust probabilities based on belief
        probabilities /= probabilities.sum()
        return np.random.choice(available_nodes, p=probabilities)

    def free_energy(belief_in_tour, path_length):
        if 0 < belief_in_tour < 1:
            uncertainty = -belief_in_tour * np.log(belief_in_tour) - (1 - belief_in_tour) * np.log(1 - belief_in_tour)
        else:
            uncertainty = 0
        expected_energy = path_length  # Use actual path length as energy
        return expected_energy + uncertainty

    for iteration in range(num_iterations):
        all_paths = []
        all_lengths = []
        for _ in range(num_ants):
            path = [random.randint(0, N - 1)]
            available_nodes = list(set(range(N)) - set(path))
            belief_in_tour = 0.5
            current_path_length = 0
            while available_nodes:
                next_node = choose_next_node(available_nodes, path[-1], belief_in_tour)
                path.append(next_node)
                available_nodes.remove(next_node)
                current_path_length += graph[path[-2], path[-1]]
                
                if best_path_length != float('inf'):
                    belief_in_tour = 1 - (current_path_length / best_path_length)
                    belief_in_tour = max(0.1, min(0.9, belief_in_tour))
            
            # Ensure the path is a complete tour by returning to the start
            path.append(path[0])
            path_length = calculate_path_length(path)
            all_paths.append(path)
            all_lengths.append(path_length)
            if path_length < best_path_length:
                best_path = path
                best_path_length = path_length

        pheromone *= (1 - evaporation_rate)
        for path, length in zip(all_paths, all_lengths):
            deposit = pheromone_deposit / length
            for i in range(len(path) - 1):
                pheromone[path[i], path[i + 1]] += deposit

        # Elitist strategy
        best_deposit = pheromone_deposit * 2 / best_path_length
        for i in range(len(best_path) - 1):
            pheromone[best_path[i], best_path[i + 1]] += best_deposit

    return best_path, best_path_length
def generate_random_graphs(num_graphs, num_nodes):
    graphs = []
    for _ in range(num_graphs):
        # Choose a random distribution type
        dist_type = np.random.choice(['uniform', 'normal', 'exponential', 'lognormal'])
        
        if dist_type == 'uniform':
            low = np.random.uniform(0.1, 1)
            high = np.random.uniform(low + 0.5, low + 5)
            graph = np.random.uniform(low, high, size=(num_nodes, num_nodes))
        elif dist_type == 'normal':
            mean = np.random.uniform(1, 5)
            std = np.random.uniform(0.1, 2)
            graph = np.abs(np.random.normal(mean, std, size=(num_nodes, num_nodes)))
        elif dist_type == 'exponential':
            scale = np.random.uniform(0.5, 2)
            graph = np.random.exponential(scale, size=(num_nodes, num_nodes))
        else:  # lognormal
            mean = np.random.uniform(0, 2)
            sigma = np.random.uniform(0.1, 1)
            graph = np.random.lognormal(mean, sigma, size=(num_nodes, num_nodes))
        
        # Ensure symmetry
        graph = (graph + graph.T) / 2
        
        # Set diagonal to zero (no self-loops)
        np.fill_diagonal(graph, 0)
        
        # Randomly set some edges to a small non-zero value (to vary density)
        mask = np.random.random(graph.shape) < np.random.uniform(0.3, 1)
        graph = np.where(mask, graph, 0.01)  # Use 0.01 instead of 0
        
        # Scale the graph to have weights mostly between 0.1 and 10
        scale_factor = 9.9 / np.percentile(graph[graph > 0], 95)
        graph = graph * scale_factor + 0.1
        
        graphs.append(graph)
    
    return graphs
    graphs = []
    for _ in range(num_graphs):
        # Choose a random distribution type
        dist_type = np.random.choice(['uniform', 'normal', 'exponential', 'lognormal'])
        
        if dist_type == 'uniform':
            low = np.random.uniform(0, 1)
            high = np.random.uniform(low + 0.5, low + 5)
            graph = np.random.uniform(low, high, size=(num_nodes, num_nodes))
        elif dist_type == 'normal':
            mean = np.random.uniform(0, 5)
            std = np.random.uniform(0.1, 2)
            graph = np.abs(np.random.normal(mean, std, size=(num_nodes, num_nodes)))
        elif dist_type == 'exponential':
            scale = np.random.uniform(0.1, 2)
            graph = np.random.exponential(scale, size=(num_nodes, num_nodes))
        else:  # lognormal
            mean = np.random.uniform(0, 2)
            sigma = np.random.uniform(0.1, 1)
            graph = np.random.lognormal(mean, sigma, size=(num_nodes, num_nodes))
        
        # Ensure symmetry
        graph = (graph + graph.T) / 2
        
        # Set diagonal to zero (no self-loops)
        np.fill_diagonal(graph, 0)
        
        # Randomly set some edges to zero (to vary density)
        mask = np.random.random(graph.shape) < np.random.uniform(0.3, 1)
        graph *= mask
        
        # Scale the graph to have weights mostly between 0 and 10
        scale_factor = 10 / np.percentile(graph[graph > 0], 95)
        graph *= scale_factor
        
        graphs.append(graph)
    
    return graphs

def compare_methods(graphs):
    results = []
    for i, graph in enumerate(graphs):
        print(f"Processing graph {i+1}/{len(graphs)}")
        
        best_path_basic, best_path_length_basic = ant_colony_optimization(graph)
        best_path_ai, best_path_length_ai = improved_aco_active_inference(graph)

        results.append({
            'graph_id': i,
            'basic_path_length': best_path_length_basic,
            'ai_path_length': best_path_length_ai,
            'improvement': (best_path_length_basic - best_path_length_ai) / best_path_length_basic * 100
        })
    return results

def analyze_graph_characteristics(graphs, results):
    characteristics = []
    
    for i, graph in enumerate(graphs):
        # Calculate graph characteristics
        total_edges = np.sum(graph > 0)
        total_possible_edges = graph.shape[0] * (graph.shape[0] - 1)
        density = total_edges / total_possible_edges if total_possible_edges > 0 else 0
        
        avg_edge_weight = np.mean(graph[graph > 0]) if total_edges > 0 else 0
        std_edge_weight = np.std(graph[graph > 0]) if total_edges > 0 else 0
        
        # Calculate the coefficient of variation of edge weights
        cv_edge_weight = std_edge_weight / avg_edge_weight if avg_edge_weight > 0 else 0
        
        # Calculate the range of edge weights
        edge_weight_range = np.max(graph) - np.min(graph[graph > 0]) if total_edges > 0 else 0
        
        characteristics.append({
            'graph_id': i,
            'avg_edge_weight': avg_edge_weight,
            'std_edge_weight': std_edge_weight,
            'cv_edge_weight': cv_edge_weight,
            'density': density,
            'edge_weight_range': edge_weight_range,
            'improvement': results[i]['improvement']
        })
    
    df = pd.DataFrame(characteristics)
    
    # Print debugging information
    print("\nDebugging Information:")
    print(f"Number of graphs: {len(graphs)}")
    print(f"Density statistics: min={df['density'].min()}, max={df['density'].max()}, mean={df['density'].mean()}, std={df['density'].std()}")
    print(f"Improvement statistics: min={df['improvement'].min()}, max={df['improvement'].max()}, mean={df['improvement'].mean()}, std={df['improvement'].std()}")
    
    # Calculate correlations
    correlations = {}
    for column in df.columns:
        if column not in ['graph_id', 'improvement']:
            correlation, p_value = spearmanr(df[column], df['improvement'])
            correlations[column] = {'correlation': correlation, 'p_value': p_value}
    
    # Print correlations
    print("\nCorrelations with improvement:")
    for char, values in correlations.items():
        print(f"{char}: correlation = {values['correlation']:.4f}, p-value = {values['p_value']:.4f}")
    
    # Plotting
    num_chars = len(correlations)
    fig, axes = plt.subplots((num_chars + 1) // 2, 2, figsize=(15, 5 * ((num_chars + 1) // 2)))
    axes = axes.ravel()
    
    for i, (column, values) in enumerate(correlations.items()):
        axes[i].scatter(df[column], df['improvement'])
        axes[i].set_xlabel(column)
        axes[i].set_ylabel('Improvement (%)')
        axes[i].set_title(f'{column} vs Improvement\nr={values["correlation"]:.2f}, p={values["p_value"]:.4f}')
    
    # Remove any unused subplots
    for j in range(i + 1, len(axes)):
        fig.delaxes(axes[j])
    
    plt.tight_layout()
    plt.show()
    
    return df, correlations
def perform_statistical_tests(results_df):
    # Paired t-test
    t_stat, p_value = stats.ttest_rel(results_df['basic_path_length'], results_df['ai_path_length'])
    print(f"Paired T-test: T-statistic = {t_stat}, P-value = {p_value}")

    # Wilcoxon signed-rank test
    w_stat, w_p_value = stats.wilcoxon(results_df['basic_path_length'], results_df['ai_path_length'])
    print(f"Wilcoxon Signed-Rank Test: W-statistic = {w_stat}, P-value = {w_p_value}")

    # Mann-Whitney U test (if treating as independent samples)
    u_stat, u_p_value = stats.mannwhitneyu(results_df['basic_path_length'], results_df['ai_path_length'])
    print(f"Mann-Whitney U Test: U-statistic = {u_stat}, P-value = {u_p_value}")

if __name__ == "__main__":
    num_graphs = 50
    num_nodes = 100

    print("Generating random graphs...")
    graphs = generate_random_graphs(num_graphs, num_nodes)

    print("Comparing methods...")
    results = compare_methods(graphs)

    results_df = pd.DataFrame(results)
    print("\nResults summary:")
    print(results_df.describe())


\end{lstlisting}

\end{document}